\pdfoutput=1
\documentclass[11pt]{article}

\usepackage{EMNLP2023}
\usepackage{times}
\usepackage{latexsym}
\usepackage{arydshln}
\usepackage{multirow}
\usepackage[T1]{fontenc}
\usepackage[utf8]{inputenc}
\usepackage{amsfonts}
\usepackage{microtype}
\usepackage{multicol}
\usepackage{inconsolata}
\usepackage{latexsym}
\usepackage{amsmath}
\usepackage{graphicx}
\usepackage{xcolor}
\usepackage{caption}
\usepackage{subfigure}
\usepackage{multirow}
\usepackage{threeparttable}
\usepackage{amssymb}
\usepackage[algo2e]{algorithm2e}
\usepackage{amsmath}
\usepackage{bbm}
\DeclareMathOperator*{\argmax}{arg\,max}

\definecolor{comment}{rgb}{0.8,0.8,0.8}
\definecolor{applegreen}{rgb}{0.55, 0.71, 0.0}
\usepackage{booktabs}
\usepackage{balance}
\usepackage{bm}
\usepackage{color}
\usepackage{colortbl}
\usepackage{colortbl}
\usepackage{textcomp}
\usepackage{CJK}
\usepackage{amsfonts}
\usepackage{microtype}
\usepackage{arydshln}
\usepackage{amsfonts}
\usepackage{enumitem}
\usepackage[T1]{fontenc}
\usepackage[utf8]{inputenc}
\usepackage{url}
\usepackage{tcolorbox}
\usepackage{xcite}
\usepackage{makecell}
\usepackage{xr}
\usepackage{pifont}
\usepackage{microtype}
\usepackage{pifont}
\usepackage{textcomp}
\usepackage{amssymb}
\usepackage{algorithm}
\usepackage{algorithmicx}
\usepackage{algpseudocode}

\usepackage{mathrsfs}
\usepackage{amsmath} %数学公式
\floatname{algorithm}{Algorithm} %算法
\title{Self-Evolution Learning for Mixup: Enhance Data Augmentation on Few-Shot Text Classification Tasks}

\author{Haoqi Zheng\textsuperscript{\rm 1}\footnotemark[1]\thanks{~~Haoqi Zheng and Qihuang Zhong contribute equally to this work.} , Qihuang Zhong\textsuperscript{\rm 2}\footnotemark[1] ,  Liang Ding\textsuperscript{\rm 3}, Zhiliang Tian\textsuperscript{\rm 1}\footnotemark[2]\thanks{~~Corresponding Authors} , \\  {\bf Xin Niu\textsuperscript{\rm 1}\footnotemark[2]} ,
{\bf Changjian Wang\textsuperscript{\rm 1}} ,
{\bf Dongsheng Li\textsuperscript{\rm 1}}, {\bf Dacheng Tao\textsuperscript{\rm 4}} \\
\fontsize{11.0pt}{\baselineskip}\selectfont  \textsuperscript{\rm 1}College of Computer, National University of Defense Technology \\
\fontsize{11.0pt}{\baselineskip}\selectfont  \textsuperscript{\rm 2}School of Computer Science, Wuhan University \;
  \textsuperscript{\rm 3}JD Explore Academy  \; \textsuperscript{\rm 4}University of Sydney
  }

\begin{document}
\maketitle
% \renewcommand{\thefootnote}{\fnsymbol{footnote}} %将脚注符号设置为fnsymbol类型，即特殊符号表示
% \footnotetext[1]{Contributed equally to this work.} %对应脚注[1]
% \footnotetext[2]{Corresponding authors.} %对应脚注[1]
% \renewcommand{\thefootnote}{\arabic{footnote}} 

\begin{abstract}
Text classification tasks often encounter few-shot scenarios with limited labeled data, and addressing data scarcity is crucial. Data augmentation with mixup merges sample pairs to generate new pseudos, which can relieve the data deficiency issue in text classification. However, the quality of pseudo-samples generated by mixup exhibits significant variations. Most of the mixup methods fail to consider the varying degree of learning difficulty in different stages of training. 
% by linearly interpolating the inputs and labels of two different samples.
% Data augmentation with mixup has shown to be effective on various text classification tasks. Mixup generates new samples by linearly interpolating the inputs and labels of two different samples.
And mixup generates new samples with one-hot labels, which encourages the model to produce a high prediction score for the correct class that is much larger than other classes, resulting in the model's over-confidence. In this paper, we propose a self-evolution learning (SE) based mixup approach for data augmentation in text classification, which can generate more adaptive and model-friendly pseudo samples for the model training. SE caters to the growth of the model learning ability and adapts to the ability when generating training samples. To alleviate the model over-confidence, we introduce an instance-specific label smoothing regularization approach, which linearly interpolates the model’s output and one-hot labels of the original samples to generate new soft labels for label mixing up. Through experimental analysis, experiments show that our SE brings consistent and significant improvements upon different mixup methods. In-depth analyses demonstrate that SE enhances the model's generalization ability.
\end{abstract}

\section{Introduction}
\label{sec:intro}
% there has been growing interest in generative large language models (LLMs)

Recently, generative large language models (LLMs) have won great popularity in natural language processing (NLP), and have achieved impressive performance on various NLP tasks~\cite{kocon2023chatgpt,peng2023towards,lu2023error}. However, empirical studies \cite{zhong2023chat} suggest that LLMs do not always outperform BERT in some language understanding tasks. Hence, employing BERT is still a viable option in some applications. Text classification tasks often encounter few shot scenarios (e.g. NLI and Paraphrase tasks), where there are limited suitable labeled data available for training. Data augmentation (DA) generates new data by changing the original data through various methods, which enlarges the training dataset to alleviate the issue of data scarcity. 
% those based on sample invariance and those based on sample equivariance. 

In text classification tasks,
DA methods can be divided into two categories: 
% the methods that only alter the inputs and the methods that modify both inputs and labels.
DA methods like EDA~\cite{wei2019eda}, Back-Translation~\cite{kobayashi2018contextual}, and others based on synthesis such as mixup. 
The first category conducts DA by altering only the inputs. These methods only alter the inputs to generate new data while maintaining the original labels. These methods are easy to implement, but the input only changes a little thus leading to augmented inputs with limited diversity,  which may reduce model generalization. The second category of DA methods modify both inputs and labels, which changes the input samples in a certain way and simultaneously changes the corresponding labels to compose a new sample.
These methods tend to generate samples more distinct from the original samples. 
% generate sample changes with greater diversity that allows the model to learn more accurate and comprehensive knowledge.

Mixup is a DA method that modifies both inputs and labels. It mixes up inputs of samples and their labels, where labels are commonly represented with one-hot encoding. Most of these methods mix up inputs of two samples on their input text  \cite{yun2019cutmix} or hidden-level representations \cite{verma2019manifold}. 
% Mixup are categorized into input-level mixup \cite{yun2019cutmix} and hidden-level mixup \cite{verma2019manifold} depending on the location of the mix operation. 
%However, mixup typically randomly selecting two samples to mix them up to compose one sample, and the generated
However, the pseudo sample, simply combined with two samples, may not be adaptive to the model's learning ability and friendly to the model training.
Recently, some work \cite{sawhney-etal-2022-dmix,park-caragea-2022-calibration} have focused on selecting similar sample pairs for the mixup.
% addressing this issue and proposing various methods for effectively choosing samples. \citet{sawhney-etal-2022-dmix} 
\citet{sawhney-etal-2022-dmix} select samples according to the embedding similarity. \citet{park-caragea-2022-calibration} merge one sample considering the confidence of the model's predictions.
% compared logits to categorize samples into two sets and interpolate samples across these two sets by finding the most similar and most dissimilar
% samples from the other set. 
% Some of these methods require the use of additional tools and are more complex, while others have overlooked the growth in the model's learning ability. 
Moreover, in few-shot scenarios, using hard labels (one-hot labels) can lead to issues, where the one-hot labels fail to provide uncertainty of inputs since all the probability mass is given to one class. This results in over-confident models since the largest logit becomes larger than the others which removes the uncertainty of label space~\cite{szegedy2016rethinking}.
% model tries to produce a logit value of the correct class that is (much larger than any of the incorrect classes, accordingly, the incorrect logits are very different from one other. It results in a too-confident model about its predictions \cite{szegedy2016rethinking}.)
The current label smoothing techniques generate soft labels that cannot dynamically adapt to the model's increasing ability as the training goes on, so they also cannot adjust according to the model's performance at the current stage.

In this paper, we propose self-evolution learning for mixup to achieve data augmentation in text classification tasks. To cater to the model's learning ability, we first divide the training data into easy-to-learn and hard-to-learn subsets. We gradually start from the mixup of easy-to-learn samples and then gradually transition to the mixup operation of hard-to-learn samples. 
% Note that the mixup is performed between similar samples. 
To avoid the model's over-confidence, we introduce an instance-specific label smoothing method, where we linearly interpolate the predicted probability distribution of the original sample and its one-hot label to obtain a soft label.  Using this soft label reduces the difference between the model's prediction probability for different classes, which can alleviate the model's over-confidence. Additionally, this instance-specific label can dynamically adapt to the growth of the model's increasing ability and can be customized to the model's current performance.
% this soft label incorporates the model's predicted probability distribution to continually adjust the label, enabling it to better adapt to the model's training. 
Our method has empirically proven that mixing up in the order of increased difficulty can make the generated samples more adaptive for model training compared to randomly selected samples.
% effective on text classification benchmarks. We prove that mixing up in the order of increased difficulty can make the generated samples more adaptive for model training compared to randomly selected samples.

% This instance-specific label can dynamically adapt to the growth of the model's increasing ability and can adjust according to the model's performance at the current stage.
% Firstly, to address the issue that the pseudo-data generated by mixup does not consider the model's learning ability, we divide the data into easy-to-learn and hard-to-learn subsets. Secondly, We start from the mixup of easy-to-learn data, and then gradually transition to the mixup operation of hard-to-learn data. Note that the mixup is performed between similar samples.

% Secondly, we gradually transition the model from easy-to-learn to hard-to-learn by performing a mixup on the easy-to-learn subset first and then training the model in increasing order of difficulty. 

% through extensive experiments 

Our contributions are as follows:

$\bullet$ We propose self-evolution learning (SE) for mixup to consider the learning difficulty of samples for data augmentations on text classification tasks. %SE  utilizes the learning difficulty of samples to generate model-friendly samples. 

$\bullet$ We propose an instance-specific label smoothing approach for regularization which can obtain dynamic and adaptive soft labels to alleviate the model's over-confidence and enhance the model's generalization ability.
% performs a linear interpolation between the model's output and the one-hot label to 

$\bullet$ Extensive experiments show that our model significantly and robustly improves the mixup method on few-shot text classification tasks.
% We have also highlighted the importance of using dynamic soft labels in the process.

\begin{figure*}[htbp]
	\centering		\includegraphics[width=\textwidth]{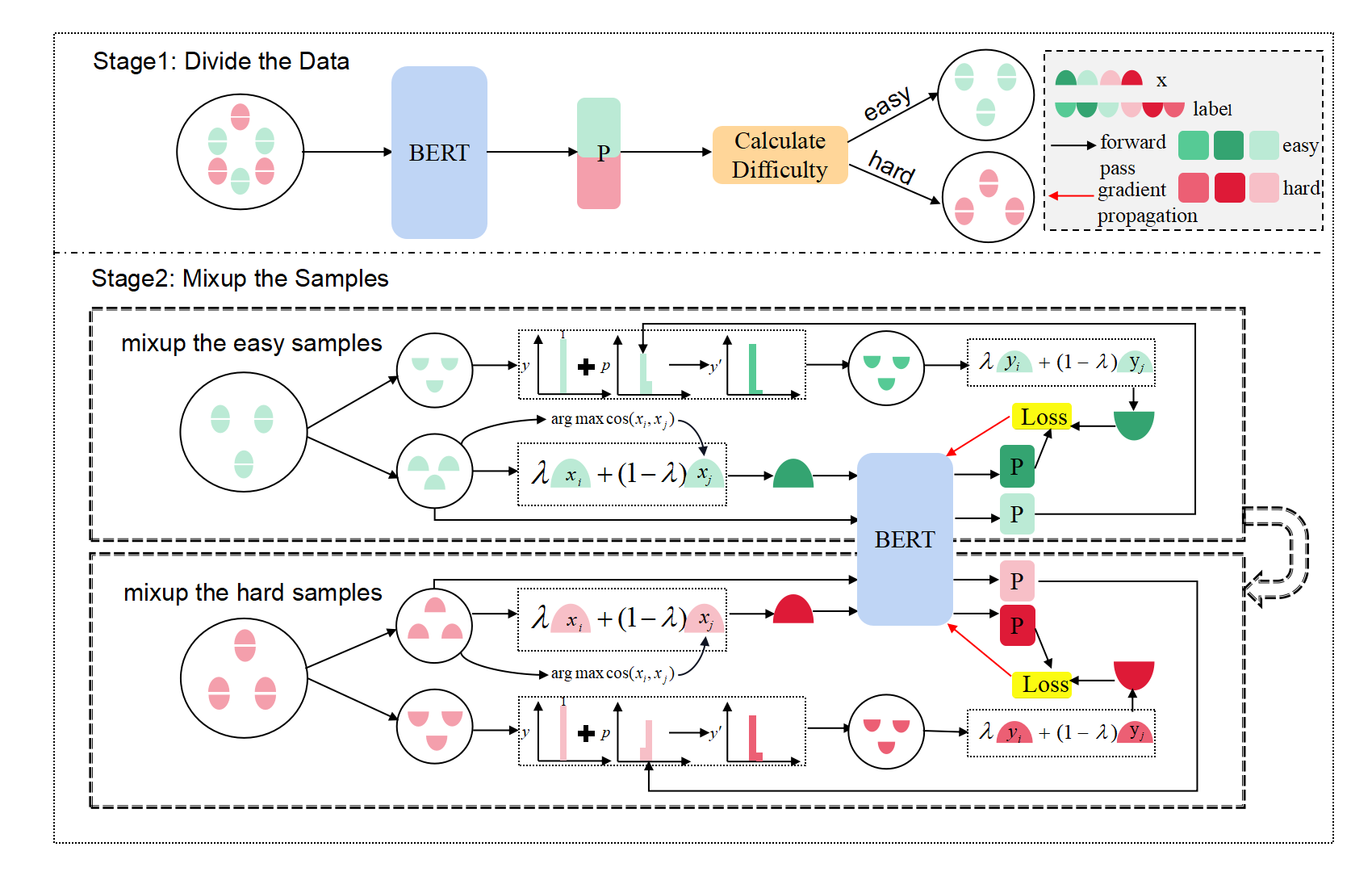}
	\centering
	\caption{: Overview of our mechanism,  which contains two stages:  using an existing BERT to divide the data according to difficulty and mixing up the samples to model's training from easy to hard. Best viewed in color.}\label{fig:overview}
\end{figure*}
\section{Related Work}
\subsection{Few-shot Text Classification}
Driven by the observation that humans can rapidly adapt existing knowledge to new concepts with limited examples, few-shot learning \cite{fei2006one} has recently drawn a lot of attention. Few-shot text classification entails performing classification after training or tuning a model on only a few examples. Several studies~\cite{yu-etal-2018-diverse,2018arXiv180402063B,geng-etal-2020-dynamic} have explored various approaches for few-shot text classification, which mainly involve the traditional machine learning techniques for selecting the optimal category sub-samples.
% \citet{yu-etal-2018-diverse} proposed an adaptive metric learning approach to dynamically select an optimal distance metric for different tasks. \citet{2018arXiv180402063B}  introduced a method for small-sample text classification using pre-trained word embeddings and manual supervision. By leveraging clustering and metric learning, they obtained category representations and assigned samples based on topic modeling principles. \citet{geng-etal-2020-dynamic} introduced the induction network, which utilizes the dynamic routing algorithm proposed by \citep{NIPS2017_2cad8fa4} to learn a generalized class-wise representation.
% Although the works mentioned above focus on the domain of few-shot scenarios, their main idea is still to improve the upper limit of model learning on known datasets. However, NLP is more concerned with cognition and understanding after perception, rather than shallow statistical semantics. Date deficiency is insufficient for the model to learn the ability of comprehension. In recent years, the emergence of pre-trained models has partially addressed the challenges of data acquisition and reuse.

More recently, ever since~\citet{devlin-etal-2019-bert,brown2020language} show the impressive performances of pre-trained language models (PLMs) on a variety of NLP tasks, a great deal of works~\cite{wu2019conditional,bansal-etal-2020-learning} tend to employ the PLMs to tackle the few-shot text classification problem. One line of work aims to fine-tune the PLMs (mainly for discriminative PLMs, such as BERT~\cite{devlin-etal-2019-bert}) with the few-shot training data. Correspondingly, how to design data augmentation methods to better enrich training data has become the focus of these works. Rather than fine-tuning the PLMs, a separate line of research aims to take full advantage of the emergent few-shot learning ability of larger PLMs, \textit{i.e.}, GPT-3~\cite{brown2020language} and InstructGPT~\cite{ouyang2022training}, and use the few-shot training data as the demonstrations for performing in-context learning process \cite{lu-etal-2023-damstf,lu2023meta}. 
% \citet{lu-etal-2023-damstf} and \citet{lu2023meta} employ meta-learning to tackle text classification tasks in few-shot scenarios.
Considering the BERT-based PLMs are more suitable for text classification tasks, we follow the former research line and focus on exploring the ability of BERT-based PLMs in the few-shot text classification.
% Pre-trained language models have also been employed in few-shot text classification. \citet{bansal-etal-2020-learning} present LEOPARD, which utilizes BERT \cite{devlin-etal-2019-bert} within an optimization-based meta-learning framework to achieve good performance across diverse NLP classification tasks. Furthermore, GPT-3 \cite{brown2020language} demonstrates that the language model itself can perform few-shot text classification without relying on meta-learning. 
% These PLMs have been adapted and fine-tuned for few-shot text classification tasks, showing improved performance compared to traditional methods. 
% Data augmentation techniques have also been explored to address the limited labeled data problem.  Our work still focuses on utilizing a pre-trained model for few-shot text classification tasks.

\subsection{Data Augmentation in NLP}
Since the bottleneck in few-shot learning is the lack of data, the performance can be easily improved if we can generate more labeled data. Hence, various NLP data augmentation techniques have been proposed, such as EDA~\cite{wei2019eda}, Back-Translation~\cite{kobayashi2018contextual} and CBERT~\cite{wu2019conditional}. These methods show remarkable performance in some specific scenarios, however, they mainly focus on altering the original input, resulting in a lack of diversity in the generated samples.
% The most commonly used method is the token replacement: randomly select tokens in a sentence and replace them with semantically similar tokens to synthesize a new sentence. \citet{wei2019eda} directly uses the WordNet \cite{miller1995wordnet} for replacement. \citet{kobayashi2018contextual} suggests employing contextual augmentation to predict the probability distribution of replacement tokens using two causal language models. \citet{wu2019conditional} introduces an extension to contextual augmentation by incorporating the masked language modeling (MLM) technique of BERT, thereby considering the bi-directional context. 
In response to this problem, 
% However, the data augmentation methods mentioned above primarily focus on altering the original input, resulting in a lack of diversity in the generated samples.
\citet{szegedy2016rethinking} first propose a domain-independent data augmentation technique (\textit{i.e.}, mixup) in the computer vision domain, that linearly interpolates image inputs on the pixel-based feature space. \citet{guo2019augmenting} then integrate the mixup with CNN and LSTM for text applications. 
Furthermore, to achieve better performance, various works~\cite{sun2020mixup,cao2021towards,chen-etal-2020-mixtext,yoon2021ssmix,zhang-etal-2022-treemix} attempt to improve the mixup technique from two perspectives: 1) how to better merge the two hidden representations, and 2) how to directly perform the mixup on the input sentences. 

% They only conduct mixup on the fixed word embedding level like \citep{szegedy2016rethinking} did in image classification. \citep{sun2020mixup} incorporates a dynamic mixup layer on top of the final hidden layer of the pre-trained transformer-based model. This mixup layer is trained together within the complete text classification model. \citet{chen-etal-2020-mixtext} proposed a mixup method named Tmix, which involves taking two different inputs, denoted as x, passing them through m layers of hidden units, and then utilizing the traditional method of mixup to merge the two hidden representations. \citet{yoon2021ssmix} employs an ingenious approach to perform mixup directly on the input of the text, instead of predominantly utilizing it on the hidden layers as seen in previous mixup methods. Under the premise of retaining the majority of important tokens, They introduce a novel span by replacing certain information, as a means of augmentation.  \citet{zhang-etal-2022-treemix} noticed that the meaning of a complex expression is built from its sub-parts and proposed TreeMix which leverages constituency parsing tree to decompose sentences into constituent sub-structures and the Mixup data augmentation technique to recombine them to generate new sentences. 
Although achieving remarkable performance, these previous mixup strategies still have some limitations. Specifically, they (usually) randomly select samples to mix and do not consider the model's learning ability. Some works~\cite{sawhney-etal-2022-dmix,park-caragea-2022-calibration} have also focused on addressing this issue and proposed various methods for effectively choosing samples. \citet{sawhney-etal-2022-dmix} select samples according to the embedding similarity.
% propose an adaptive distance-aware interpolative mixup that selects samples based on their diversity in the embedding space. 
\citet{park-caragea-2022-calibration} merge one sample considering the confidence of the model's predictions.
% compare logits to categorize samples into two sets and interpolate samples across these two sets by finding the most similar and most dissimilar samples from the other set. 
Along the same research line, in this paper, we improve the mixup with a more simple-yet-effective self-evolution learning mechanism. 

%  First,    
% Second, most of these mixup methods generate samples with one-hot labels. Along the same research line, in this paper, we improve the mixup with a self-evolution learning mechanism. 

% \section{Engines}

% \section{Preamble}

\section{Method}

\subsection{Overview}
For the text classification task in the few-shot scenario, we propose a data augmentation method via mixup, where the training follows an easy-to-hard schedule over the augmented data. First, we construct a  text classification model based on the BERT and then employ a mixup method for data augmentation to expand the amount of data (Sec\ref{sec:4.2}). To make the mixup adaptive for model learning ability, we propose \textbf{self-evolution learning} for mixup (Sec\ref{sec:4.3}). To alleviate the over-confidence problem of the model, we propose an \textbf{instance-specific label smoothing regularization method}, which linearly interpolates the model's outputs and one-hot labels of the original samples to generate new soft labels as the label for mixing up (Sec\ref{sec:4.4}). 
% In this paper, we utilize a pre-trained BERT model for text classification tasks, and in the few-shot scenario, we employ the mixup data augmentation method by linearly interpolating labeled data pairs to generate new samples and enhance the classification performance of the BERT model. 

% Tan better adapts the generated mixup data to the model's ability to absorb samples, and we propose a self-evolution learning for the mixup algorithm. The algorithm trains mix up samples in an easy-to-hard manner by partitioning the original training set based on the difficulty level and gradually performing mixup training from easy to hard.

% To mitigate the issue of over-confidence caused by adopting hard labels in mixup, we propose an instance-specific label smoothing(\textbf{ILS}) regularization method, which linearly interpolates the model's output and one-hot label of the original samples to generate new soft labels that are used for label mixing in the mixup.

\subsection{Text Classification Model and Mixup}\label{sec:4.2}
We utilize the BERT \cite{2018arXiv181004805D} for text classification tasks, where the BERT model adopts a multi-layer bidirectional Transformer encoder architecture and is pre-trained on plain text for masked language modeling. 

BERT takes a sequence of words as the input and outputs the representation of the sequence.
% The sequence has one or two segments the first token of the sequence is always $[CLS]$ which contains the special classification embedding
% and another special token $[SEP]$ is used for separating segments. 
For text classification tasks, BERT takes the final hidden state $h$ of the first token $[CLS]$ as the sentence representation. Then, we append a softmax function with a linear transformation to generate a probability distribution and the predicted label.
 %BERT is a language representation model, which stands for Bidirectional Encoder Representations from Transformers\cite{2018arXiv181004805D}.BERT is mainly composed of an embedding layer, encoder layer, and pooling layer. 
 % For a sentence S, the embedding layer converts the tokenized S into a vector representation. The encoder layer is responsible for extracting text features and effectively learning contextual information. The pooling layer aggregates the outputs of the encoder layer into a fixed-size representation of the entire sequence. For text classification tasks, an additional output layer is added based on the pooling layer. The output layer uses a fully connected layer and the softmax function to obtain the predicted probabilities for each category, which is used for text classification.
 
To relieve the data deficiency in few-shot scenarios, we propose a data augmentation method to generate pseudo samples for training the BERT model. The core idea of mixup is to select two labeled data points (${x}_i,{y}_i$) and (${x}_j,{y}_j$), where ${x}$ is the input and $y$ is the label. The algorithm then produces a new sample ($\tilde{x}$, $\tilde{y}$) through linear interpolation:
\begin{gather}
{
\tilde{x}=\lambda x_i + (1-\lambda) x_j \label{eq:1}} \\
{\tilde{y}=\lambda y_i + (1-\lambda) y_j \label{eq:2}
} 
\end{gather}
where $\lambda \in [0,1] $  denotes the mixing ratio of two samples. 

%%%%%%%%%%%%%%%%%%%%% Algorithm %%%%%%%%%%%%%%%%%%%%%%%%%%5
\begin{algorithm}[t]
	%\SetAlgoVlined
	\DontPrintSemicolon
\textbf{Input}:
Labeled set $ \mathcal{D} $;
%Beta distribution parameter $\alpha$;
Mixup function $\zeta(\cdot)$; 
Instance-specific label smoothing function $\phi(\cdot)$
%Discriminator function $d(\cdot)$; Number of expected generation $N$.

// \textbf{Stage1: Divide the Data} \\
// \textbf{Calculate Difficulty} \\
\For {$( {x_i, y_i} )$ $\mathbf{in}$ $\mathcal{D}$}
{
  $ {d}(x_i) = 1-{(p(y_i|x_i)-\max_{y \in C, y \neq y_i} p(y|x_i)  )}$
   % {$\mathbf{d_i} = \mathbf{pred_{yi}-\max}\left\langle \mathbf{pred_{j , j\neq yi}}  \right\rangle $}
}
 $\xi= \textbf{median}({d(x_i)})$ \\
\For {$( {x_i, y_i} )$ $\mathbf{in}$ $\mathcal{D}$}
{
     $\mathcal{D}_{easy} \leftarrow \mathcal{D}_{easy} \cup ( {x}_i, {y}_i ), \quad  if \quad d(x_i) \leq \xi $

      $\mathcal{D}_{hard} \leftarrow \mathcal{D}_{hard} \cup ( {x}_i, {y}_i ), \quad if \quad d(x_i) > \xi $
    % \If{$d_i \leq judge$}
    %    {\mathcal{D_{easy}}=$\mathcal{D_{easy}} \cup \mathbf{x_i}$}\\
    % \EndIf
    % \If{$d_i \geq judge$}  
    %    {\mathcal{D_{easy}}=$\mathcal{D_{easy}} \cup \mathbf{x_i}$}
     % \If{$d_i \geq judge$}
     %            {
     %                $\mathcal{D}_{easy} \leftarrow \mathcal{D}_{easy} \cup ( \mathbf{x}_i, \mathbf{y}_i )$
     %            }
                
     %          \If{$d_i \leq judge$}
     %          {
     %           $\mathcal{D}_{hard} \leftarrow \mathcal{D}_{hard} \cup ( \mathbf{x}_i, \mathbf{y}_i )$
     %          }
}
// \textbf{Stage2: Mixup the Samples} \\
\For {$( {x_i, y_i} )$ $\mathbf{in}$ $\mathcal{D}_{easy}$}
{
 ${x}_j={x}_{\argmax \cos{( {x_i},{x_j} )} }$\\
  $y'_i=\phi(y_i) $\\
  $y'_j=\phi(y_j)$\\
$\tilde{{x}_i}=\zeta({x_i},{x_j})$\\
$\tilde{{y}_i}=\zeta({y'_i},{y'_j})$
 
}

\For {$( {x_i, y_i} $ $\mathbf{in}$ $\mathcal{D}_{hard}$}
{
${x}_j={x}_{\argmax \cos{( {x_i},{x_j} )} }$\\
  $y'_i=\phi(y_i) $\\
  $y'_j=\phi(y_j)$\\
$\tilde{{x}_i}=\zeta({x_i},{x_j})$\\
$\tilde{{y}_i}=\zeta({y'_i},{y'_j})$
}

\KwOut{$ \{ (\tilde{{x}_i}, \tilde{{y}_i} ) \}_{i=1}^m $}
\caption{Mixup with Self-evolution learning and instance-specific label smoothing}
\label{algo:seqmix}
\end{algorithm}
%%%%%%%%%%%%%%%%%%%%%%% Algorithm %%%%%%%%%%%%%%%%%%%%%%%%

% \renewcommand{\thealgorithm}{2} %
%     \begin{algorithm}
%         \caption{Self-evolution learning} 
%         \begin{algorithmic}[1] 
%             \Require $D$ = \{$x_1,x_2,...,x_N$$M$\{$u_1$,$u_2$\
%             \Ensure \{$C_1$,$C_2$\}
%             \For{$m = 1 \to M$} //$m$
%                 \State $C_1 \Leftarrow \emptyset, C_2 \Leftarrow \emptyset$ //
%                 \For{$i = 1,2,...,N$}     //$i$
%                     \State $d_{i1} \Leftarrow {\Vert x_i-u_1 \Vert}^2$, $d_{i2} \Leftarrow {\Vert x_i-u_2 \Vert}^2$ //
%                     \If {$d_{i1} \leq d_{i2}$}
%                     \State $C_1 \Leftarrow C_1 \cup \{x_i\}$ //
%                     \Else
%                     \State $C_2 \Leftarrow C_2 \cup \{x_i\}$ 
%                 \EndIf
%                 \EndFor
%                 \State $\tilde{u_1} \Leftarrow \frac{1}{\vert C_1 \vert}\sum_{x \in C_1} x$, $\tilde{u_2} \Leftarrow \frac{1}{\vert C_2 \vert}\sum_{x \in C_2} x$ //
%                  \If{$(\tilde{u_1} == u_1)  (\tilde{u_2} == u_2$}
%                     \State \textbf{break} from line 3 %\textbf
%                     \Else
%                     \State $u_1 \Leftarrow \tilde{u_1}, u_2 \Leftarrow \tilde{u_2}$ //
%                 \EndIf
%             \EndFor
%             \State \Return $C_1, C_2$ //
%         \end{algorithmic}
%     \end{algorithm}

\subsection{Self-Evolution Learning for Mixup}\label{sec:4.3}

To make the mixed samples more adaptive and friendly to the model training, we propose a novel mixup training strategy: progressive mixup training from easy to hard. This idea is inspired by human learning behavior: a human's learning schedule usually starts from easier tasks and gradually progresses to more challenging tasks. We first propose the degree of \textit{difficulty} to measure the difficulty of the model in learning samples and then conduct mixup in two stages: (1) dividing the dataset based on the degree of difficulty, and (2) mixup two samples according to the order of difficulty from easy to hard.  

To obtain the degree of difficulty $d(x_i)$ for sample $x_i$, we
calculate the difference between the model predicted probability on the correct label $p(y_i|x_i)$ and the maximum predicted probability among the wrong labels as Eq.\ref{eq:3}:
% Specifically, for a given sample ($x_i,y_i$), we use the correct label predicted probability minus the maximum wrong label predicted probability, then we can obtain the degree of difficulty $d_i$ as Eq.\ref{eq:3}
\begin{gather}
{
{d}(x_i) = 1-(p(y_i|x_i)-\max_{y \in C, y \neq y_i} p(y|x_i)), \label{eq:3}
 }
\end{gather}
where $y_i$ denotes the ground-truth label, and $C$ denotes the set of all candidate labels.
% The $p(y|x_i)$ can reflect the model's confidence in its predictions for the sample. Higher $p(y_i|x_i)$ indicates the lower difficulty of the sample for the model.

In the first stage of self-evolution learning (SE), we divide the training data into two datasets according to the degree of difficulty. Given a training set $D$, we calculate the degree of difficulty of each sample as mentioned in Eq.\ref{eq:3}. Then, we use the median of the degree of difficulty to partition the dataset: we assign samples with a degree of difficulty less than the median to the easy-to-learn
%修改
dataset $D_{easy}$, and samples with the degree of difficulty greater than the median to the hard-to-learn dataset $D_{hard}$.

In the second stage of self-evolution learning, we conduct mixup from $D_{easy}$ to $D_{hard}$. For easy-to-learn data, we perform mixup operations on the $D_{easy}$. Given a sample $x_i$ from $D_{easy}$, we search for the most similar sample $x_j$ in $D_{easy}$, where the similarity is measured by cosine similarity. Then, we mix the two samples up by interpolating the inputs ($x_i$ and $x_j$) and labels ($y_i$ and $y_j$) as Eq.\ref{eq:1} and Eq.\ref{eq:2}.  The data selected according to the above process is then used for training, and the resulting generated data is added to the model training. In the hard-to-learn dataset, we follow the same way that selects two most similar samples and mixup to compose a pseudo sample. The sample serves as a new sample to augment the training data. Algorithm 1 summarizes the above procedure.

\subsection{Instance-Specific Label Smoothing for Regularization} \label{sec:4.4}
To avoid over-confidence caused by hard labels in few-shot scenarios, we propose a novel instance-specific label smoothing (ILS) approach to adaptively regularize the training and improve the generalization ability of the classification model.

The traditional label smoothing (LS) approach replaces the hard label distribution $y_i$ with $y'_i$ as Eq.~\ref{eq:5}, where $y'_i$ is a mixture of the original label distribution $y_i$ and a distribution $u_i$. The $u_i$ is usually a uniform distribution.

\begin{gather}
  { y'_i=(1-\alpha)*y_i+\alpha u_i} 
  \label{eq:5}
\end{gather}

% that is usually a uniform distribution, with weights $1-\alpha$ and $\alpha$. 

Traditional LS lowers the value of the correct label and increases all others, which successfully prevents the largest predicted score much larger than all others \cite{szegedy2016rethinking}. However, in the traditional LS, the distribution of $u$ is fixed and $u$ cannot dynamically generate labels to adapt to the model learning. 
% hardly provides enough linguistic information to guide the model training.
% LS approach aims to minimize the cross-entropy between the soft label $y'_i$ and the predicted output $p_i$ of the model formulated as:
% \begin{gather}
%   { y'_i=(1-\alpha)*y_i+\alpha u_i} 
% \end{gather}
% where $u_i$ is a fixed distribution that is usually a uniform distribution, and $\alpha$ is a weighting factor.
% Mathematically, in LS approach, minimizes the cross-entropy between modified label distribution $y'_i$ and the model output $p_i$, where $y'_i$  is the smoothed label distribution formulated as:

% Additionally, based on \cite{YuanTLWF20}, we reformulate loss function for LS as:
% \begin{gather}
%   { \mathcal{L}_{LS}=(1-\alpha)*H(y,p)+\alpha D_{kl}(u,p)} 
%   \label{eq:5}
% \end{gather}
% where $H$ denotes the ordinary cross-entropy loss and $D_{kl}$ denotes the KL divergence loss. 

% We can regard $D_{kl}(u, p)$ as a knowledge distillation process, where $u$ represents a virtual teacher that guides the training of the classification model. However, the $u$ cannot follow the improvement of the student model and hardly provides enough linguistic information to guide the training of the model.

Motivated by this observation, in our instance-specific label smoothing, we propose a sample-aware prior distribution to smooth the labels. Specifically, we replace the fixed distribution $u$ with a dynamic and informative distribution that is adaptively generated by the classification model itself. In practice, similar to Eq.~\ref{eq:5}, we smooth the label by interpolating the original label $y_i$ with a $p(y|x_i)$ predicted by the classification model. Over all the candidate classes $y_i$ is a one-hot vector, where its value (i.e. probability) on the correct class is 1 and its value on the other class is 0. $p(y|x_i)$ is the model's predicted probability distribution over all the classes. We consider the model prediction $p(y|x_i)$ as the possibility of being the correct label from the model's perspective. As the model is optimized, the model prediction becomes increasingly accurate and the model predicted label approaches the ideal label. We obtain the final smoothed label ${y'}_i$ as:
%in addition to its original one-hot original label $y_i$ with a  label $y_i$, we introduce a the model's predicted probability distribution $p(y|x_i)$ over all the classes to as a degree reflecting the reference probabilities $r_i$. Then, similar to Eq.\ref{eq:5}, we can obtain the smoothed label $\tilde{y}_i$ via interpolation:
\begin{gather}
  { {y'}_i=(1-\alpha)*y_i+\alpha r_i} 
\end{gather}

Then we get the mixed smooth label $\tilde{y}'_i$ through the Eq. \ref{eq:2}. 
% \begin{gather}
%   { \tilde{y'}_i=(1-\alpha)*y_i+\alpha r_i} 
% \end{gather}
%Incorporating the model's predictive probability distribution into the label distribution enables the model to learn cautiously from difficult samples and better capture information from them, mitigating the overconfidence problem.

Finally, in the SE training stage, we employ the cross-entropy loss as follows:
% Lastly, we use the cross-entropy as the loss function in the SE training stage, as follows:
\begin{gather}
  { \mathcal{L}_{LS}=-\frac{1}{m} \sum_{i=1}^m \tilde{y}'_i \log p_i } 
\end{gather}

\begin{table*}[]
\centering
\scalebox{0.9}{
    \begin{tabular}{lccccccccc}
    \toprule
    \multicolumn{1}{c}{}                         &\bf SST2                        &\bf RTE                                             &\bf MRPC                         &\bf CB         &\bf Rott.            &\bf SUBJ           &\bf Amazon          & \multicolumn{2}{c}{\bf Score}      \\
    \cmidrule(lr){2-8} \cmidrule(lr){9-10} 
    \multicolumn{1}{l}{\multirow{-2}{*}{\bf Method}} & \textit{Acc.}               & \textit{Acc.}                             & \textit{Acc.}               & \textit{Acc.} & \textit{Acc.} & \textit{Acc.} & \textit{Acc.} & \textit{\underline{Avg.}} & \textit{$\Delta$ ($\uparrow$)} \\
    \midrule \midrule
    \multicolumn{10}{l}{\textit{Performance of Different No Mixup Methods }}  \\ \midrule
    BERT-base                               &55.43 & 49.59 & 53.67 & 32.49 & 59.46& 82.72 & 56.46 & \underline{55.68 }          & --              \\
    \cdashline{1-10}
    \quad -w/ EDA &54.72 &49.80 &59.20 &39.60 & 58.96 & 82.06 & 62.2 & \underline{58.07}         & \textcolor[RGB]{0,176,80}{+2.39}             \\
    \quad -w/ Back Translation & 56.85 &49.20&61.60 & 38.94& 60.67&83.16 &66.63 & \underline{59.57}        & \textcolor[RGB]{0,176,80}{+3.89}             \\
    \quad -w/ CBERT &54.90 &49.80 &57.20 & 32.85& 60.03&82.49 &62.82 & \underline{57.15}           & \textcolor[RGB]{0,176,80}{+1.47}             \\
    \midrule
    \multicolumn{10}{l}{\textit{Performance of Different TMix Improvement Methods }}  \\ \midrule
    TMix                 &            54.94 & 49.60 & 61.90 & 41.06 &	56.95 &	83.16 & 58.14  & \underline{57.95 }          & --              \\
    \cdashline{1-10}
    \quad -w/ AUM &56.60 & 49.81 & 62.10 & 42.35 &58.94 & 83.30 & 65.22 & \underline{59.75}        & \textcolor[RGB]{0,176,80}{+1.80}             \\
    \quad -w/ DMix & 53.68 & 54.40 & 46.40 & 56.80 & 41.80 & 51.76 & 88.66 & \underline{56.21}        & \textcolor[RGB]{255,0,0}{-1.74}             \\
    \quad -w/ SE (Ours) &57.56 &49.99 &62.69& 42.85& 58.23 &83.87 &68.58 & \underline{60.53}           & \textcolor[RGB]{0,176,80}{+2.58}             \\
     \midrule
    \multicolumn{10}{l}{\textit{Performance upon Different Mixup Methods }}  \\ \midrule
     SSMix  &    55.70 & 49.52 & 60.10 & 37.13 &	59.86&	83.76 & 62.63 &  \underline{58.08 }          & --              \\
     \quad -w/ SE (Ours) &56.96 &49.96 &61.41& 39.63& 61.27 &84.06 &65.60 & \underline{59.83}          & \textcolor[RGB]{0,176,80}{+1.45}             \\
    \cdashline{1-10}
    EMbedMix  & 53.11 & 49.52 & 61.61 & 37.49 &	58.83 &	83.10 & 63.34 &  \underline{58.14}          & --              \\
     \quad -w/ SE (Ours) &55.89 &49.88 & 63.28 & 41.07 & 60.10 & 83.86 &69.22 & \underline{60.46}           & \textcolor[RGB]{0,176,80}{+2.32} \\
        \cdashline{1-10}
         TreeMix  & 55.70 & 49.52 & 60.04 & 37.13 &	59.86&	83.76 & 62.63 &   \underline{58.37}          & --              \\
     \quad -w/ SE (Ours) &56.96 &49.96 & 61.17 & 39.63 & 61.27 & 84.06 &65.60 & \underline{59.80}          & \textcolor[RGB]{0,176,80}{+1.43}  \\
    
\bottomrule    
\end{tabular}
}
\caption{Comparison between our SE and the vanilla method applied to mixup methods on the benchmarks. ``Rott.'' is the short for Rotten tomato task. ``$\Delta$'' denotes the improvement of SE methods compared to the baselines. 
% It shows that SE method yields consistent improvements among all PLMs, indicating the effectiveness and universality of SE method.
}
\label{tab:main1}
\end{table*}

\begin{table}
\centering
\scalebox{0.8}{
\begin{tabular}{lccccc}
\toprule
\textbf{Model} & \textbf{CB} & \textbf{RTE}  & \textbf{Rott.} & \textit{Avg.} & \textit{$\Delta$ ($\uparrow$)}\\
% \cline{2-4}
\midrule
Baseline & 37.84 & 48.51 & 58.55 & 48.30 &{--}\\
\cdashline{1-6}
\quad -w/ SSMix &  42.49 & 48.37 &  59.67 &50.17 &  \textcolor[RGB]{0,176,80}{+1.87}        \\
\quad -w/ SE (Ours) &\makecell{\textbf{47.49}%\\ (+9.65)\\ \textbf{(+5.00)}
} & \makecell{\textbf{49.16}%\\(+0.65)\\ \textbf{(+0.89)}
} & \makecell{\textbf{62.26}%\\(+3.71)\\ \textbf{(+2.59)}
}    & \textbf{52.97} &  \textcolor[RGB]{0,176,80}{+4.67}     \\
% BERT-large+SSMix & 42.49 & 48.37 &  59.67 \\
% \hline
% BERT-large+SSMix+SE & \makecell{47.49%\\ (+9.65)\\ \textbf{(+5.00)}
% } & \makecell{49.16%\\(+0.65)\\ \textbf{(+0.89)}
% } & \makecell{62.26%\\(+3.71)\\ \textbf{(+2.59)}
% } \\
% \citeyearpar{ct1965} & \verb|\citeyearpar| & \verb|\shortcite| \\
% \citeposs{ct1965} & \verb|\citeposs| & no equivalent \\
% \citep[FFT;][]{ct1965} &  \verb|\citep[FFT;][]| & no equivalent\\
\bottomrule
\end{tabular}}
\caption{\label{t-3}
Experimental results of comparison with BERT-large.All values are average accuracy (\%) of five runs with different seeds. Models are trained with 10 labeled data per class. 
}
\end{table}

\section{Experiments}

\subsection{Datasets}
% As presented in Table \ref{tab:dataset_summary}, we conducted experiments on various text classification benchmarks to evaluate the effectiveness of our method. We randomly selected 10 samples per class from each dataset to form the training set for training.
To investigate the effectiveness of our method, we conduct extensive experiments on various language understanding tasks, including a diversity of tasks from GLUE~\cite{wang2018glue}, SuperGLUE~\cite{wang2019superglue} and other benchmarks, \textit{i.e.}, sentiment analysis (SST-2, Rotten tomato), natural language inference (RTE, CB), paraphrase (MRPC), and text classification (SUBJ, Amazon counterfactual). To simulate the few-shot scenarios, we randomly select 10 samples per class from the training set for each task, and use them for training the models. 
For evaluation, we use the Accuracy as the metric and report the averaged results over 5 random seeds to avoid stochasticity.
Due to the space limitation, we show the details of all tasks and datasets in Appendix~\ref{appendix_data} (Table~\ref{tab:dataset_summary}).

\subsection{Implementation Details}
% We use BERT-base-uncased for the text classification task as the underlying model. We perform all experiments with five different seeds and report the average score. We set a maximum sequence length of 128, and a batch size of 32, with an AdamW optimizer with a weight decay of 1e-4. We use a linear scheduler with a warmup for 10\% of the total training step. We update the best checkpoint by measuring validation accuracy on every epoch.
We use the representative BERT~\cite{devlin-etal-2019-bert}-\textsc{Base} and -\textsc{Large} models as the backbone PLMs, and fine-tune them in a two-stage manner. Specifically, following many previous mixup methods~\cite{chen-etal-2020-mixtext,yoon2021ssmix}, we first train the backbone PLMs (without using mixup) with a learning rate of 5e-5, and then continue fine-tuning the models using the mixup strategy with a learning rate of 1e-5. Note that our methods are only adopted in the second stage. 

We set a maximum sequence length of 128 and a batch size of 32. AdamW~\cite{loshchilov2018decoupled} optimizer with a weight decay of 1e-4 is used to optimize the model. We use a linear scheduler with a warmup for 10\% of the total training step.

% We conduct training in two stages: we first train without mixup with a learning rate of 5e-5, and then train with mixup starting from the previous training’s best checkpoint, with a learning rate of 1e-5.

% For self-evolution learning, we use our method on various mixup methods. During mixup training, for each iteration, we start with mixup on the easy-to-learn dataset and then mixup on the hard-to-learn dataset. For instance-specific label smoothing, we compared the prior ratio $\lambda$ of label smoothing from 0.1 to 0.9 and found out that setting the  $\lambda$  to 0.1 is the optimal hyperparameter.

\subsection{Compared methods}
We compare our method with other cutting-edge counterparts. Specifically, taking the TMix~\cite{chen-etal-2020-mixtext} as the base mixup method, we use the following strategies to improve its performance:
\begin{itemize}
    \item AUM \cite{park-caragea-2022-calibration}: AUM compares logits to classify samples into two sets and then interpolates samples between these sets by identifying the most similar and most dissimilar samples from the opposite set.
    \item DMix \cite{sawhney-etal-2022-dmix}: DMix chooses samples based on their diversity in the embedding space.
    \item SE (Ours): SE divides the dataset into easy-to-learn and hard-to-learn and then mixes up two samples according to the order of difficulty from easy to hard.
\end{itemize}

Additionally, for reference, we report the results of some traditional data augmentation methods, \textit{i.e.,} EDA~\cite{wei2019eda} , Back Translation~\cite{shleifer2019low} and CBERT~\cite{wu2019conditional}. To verify the universality of our SE, we also attempt to adopt it to other base mixup methods, \textit{i.e.}, EmbedMix~\cite{guo2019augmenting}, SSMix~\cite{yoon2021ssmix} and TreeMix~\cite{zhang-etal-2022-treemix}.

% \textbf{BERT} \cite{2018arXiv181004805D} is a pre-trained model. We use the BERT-based-uncased model for the text classification tasks. Notably, the BERT model did not use the data augmentation technique.

% \textbf{EmbedMix} is a method that mixup on the embedding layer, which is similar to \citealp{2019arXiv190508941G}.

% \textbf{TMix} \cite{chen-etal-2020-mixtext}  is a mixup technique that interpolates the hidden states of two distinct inputs at a particular encoder layer and subsequently feeds the combined hidden states forward to the remaining layers.

% \textbf{SSMix} \cite{2021arXiv210608062Y} is a mixup method where the operation is performed on the input text and uses span-based mixing to synthesize a sentence.

\subsection{Main Results}
% Table \ref{t-2} illustrates our main results. From the result, it is clear that our method outperforms the original mixup methods in various tasks and demonstrates a substantial improvement in low-resource scenarios. 
 % We also observe that mixup avails improvement effects on BERT over multiple tasks. 
The full results of BERT-\textsc{Base} and -\textsc{Large} are shown in Table~\ref{tab:main1} and Table~\ref{t-3}, and we can find that:

\paragraph{SE surpasses the cutting-edge counterparts in most settings. } 
When using the TMix as base method, our SE brings much better performance improvements compared to the other counterparts (AUM and DMix), \textit{i.e.}, up to +4.51 average score. Additionally, compared to the other traditional DA methods, SE can also achieve superior performance. These results show the effectiveness of our SE method.

\begin{table}
\centering
\scalebox{0.83}{
\begin{tabular}{lcccc}
\toprule
\textbf{Learning Strategy} & \textbf{SST2} & \textbf{Rott.} & \textbf{Amazon} & \textit{Avg.}\\
\midrule
Random & 55.70 & 59.86 & 60.12 &58.56\\
Easy-to-hard & \textbf{55.81} & \textbf{61.17} & \textbf{65.37} & \textbf{60.78}\\
Hard-to-easy &  55.79 & 61.13 & 64.64 & 60.52\\
\bottomrule
\end{tabular}}
\caption{\label{t-5}
Experimental results of different data selection methods. All values are average accuracy (\%) of five runs with different seeds. Models are trained with 10 labeled data per class. 
}
\end{table}

% In comparison to other no-mixup data augmentation methods, SE achieves superior performance across multiple tasks. 
% When compared to AUM and DMix, while utilizing the same foundation of TMix, SE exhibits greater enhancements in the majority of tasks. It is evident that our proposed SE approach proves highly effective in enhancing the mixup strategy

\paragraph{SE brings consistent and significant performance gains among all baselines.} In addition to the TMix, we also adopt our SE to more base mixup methods, \textit{i.e.}, SSMix, EmbedMix and TreeMix, and show the contrastive results in Table~\ref{tab:main1}. As seen, compared to the baselines, our SE can bring consistent and significant performance gains among all these methods, indicating its universality.

% The incorporation of our SE method into various mixup approaches has led to improvements in performance across multiple tasks. This demonstrates the effectiveness and scalability of our proposed strategy for enhancing mixup.

\paragraph{SE works well in both model sizes.}
Here, we verify whether our SE can still work in the large model scenarios. Taking some tasks as examples, we show the contrastive results in Table~\ref{t-3}. It can be seen that, with the help of our SE, BERT-large achieves much better performance against the baselines. These results prove the effectiveness of our SE in both model sizes.
% Through the results of Table 1 and Table 2, we can see that SE has achieved significant improvements on both BERT-\textsc{BASE} and BERT-\textsc{LARGE}.

\begin{figure}
\centering %表示居中
\includegraphics[width=0.45\textwidth]{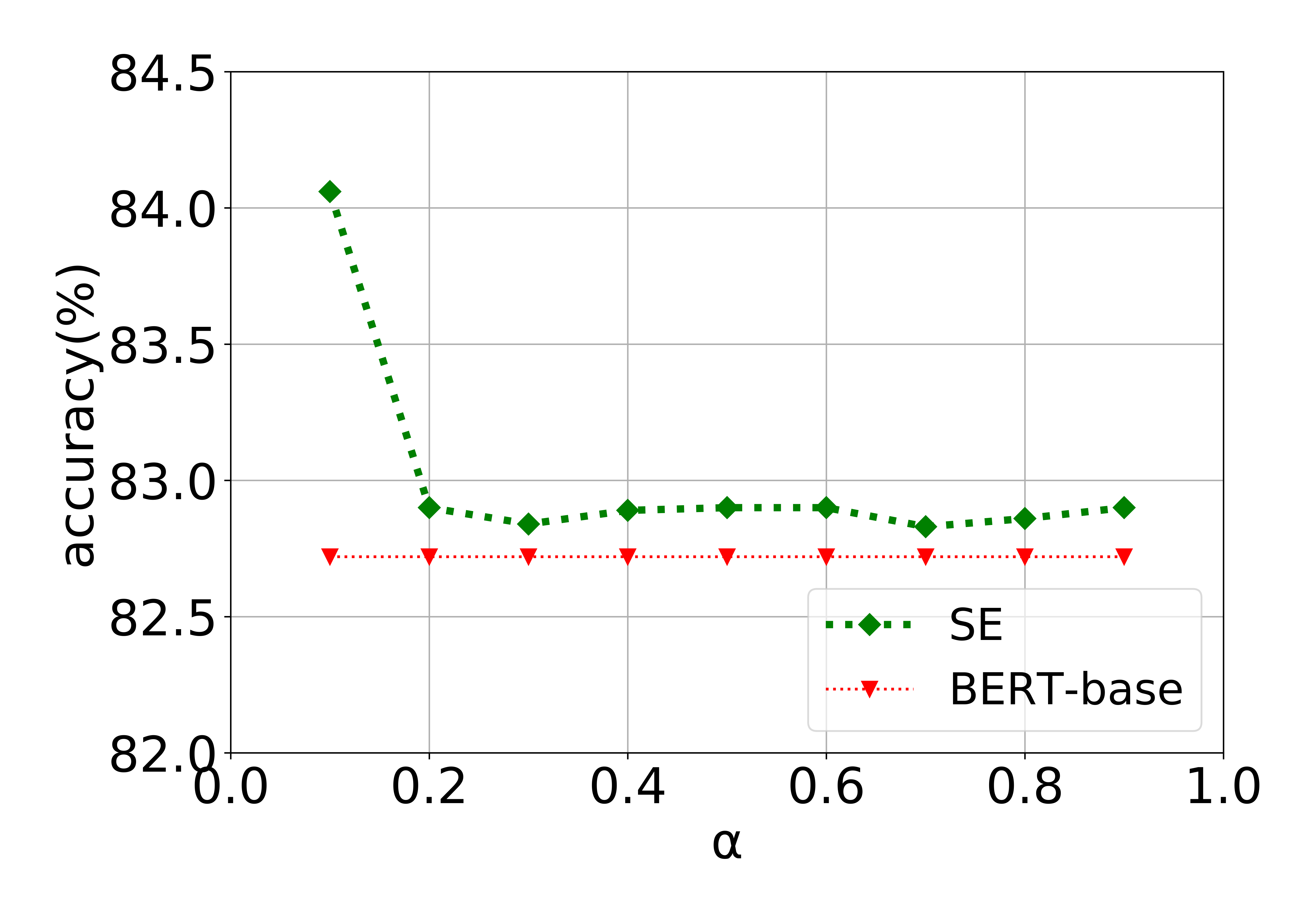}
%[height=4.5cm]表示高度
%[width=9.5cm]表示宽度
%{111.eps}表示eps格式的图片，名为111
\caption{ Parameter analysis of $\alpha$ on BERT-base, fine-tuned on SUBJ task.}
\label{fig:subfig2}
\end{figure}

\begin{table}[t]
\centering
\resizebox{\linewidth}{!}{
\begin{tabular}{lccccc}
\toprule
\textbf{Method} & \textbf{SST2} & \textbf{RTE} & \textbf{Amazon} & \textit{Avg.} & \textit{$\Delta$ ($\uparrow$)} \\
\midrule
SSMix& 55.81&49.73 &65.37 & 56.97 & {--}\\
\cdashline{1-6}
  \quad {-}w/ Vanilla LS & 56.12 & 49.81 & 65.11  & 57.01 & \textcolor[RGB]{0,176,80}{+0.04}\\
  % \hline
\quad {-}w/ ILS (Ours) & \textbf{56.88} & \textbf{49.96} & \textbf{65.52}& \textbf{57.45} & \textcolor[RGB]{0,176,80}{+0.48}\\
% \citeyearpar{ct1965} & \verb|\citeyearpar| & \verb|\shortcite| \\
% \citeposs{ct1965} & \verb|\citeposs| & no equivalent \\
% \citep[FFT;][]{ct1965} &  \verb|\citep[FFT;][]| & no equivalent\\
\bottomrule
\end{tabular}}
\caption{\label{t-6}
Experimental results of different label smoothing. All values are average accuracy (\%) of five runs with different seeds. Models are trained with 10 labeled data per class. 
}
\end{table}

\begin{figure}[t]
\centering %表示居中
\includegraphics[width=0.48\textwidth]{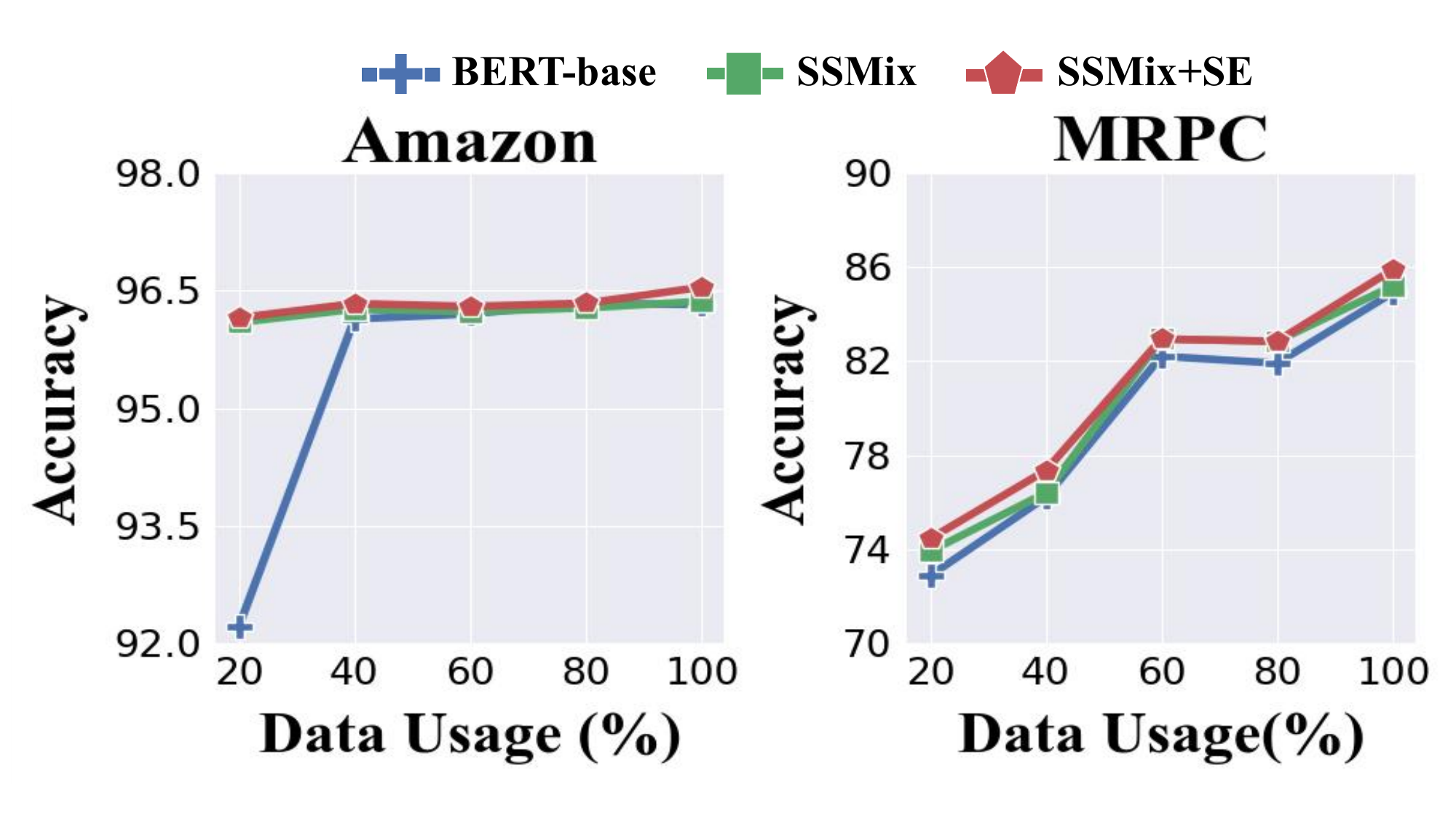}
%[height=4.5cm]表示高度
%[width=9.5cm]表示宽度
%{111.eps}表示eps格式的图片，名为111
\caption{Results at various training data sampling rates. BERT-base models fine-tuned on Amazon and MRPC are used. We can see that our method achieves better performance across all data size regimes, especially in the few-shot scenarios.}
\label{f1}
\end{figure}
% \paragraph{Impact of the Number of Labeled Data.} 

\begin{figure}[t]
\centering %表示居中
\includegraphics[width=0.48\textwidth]{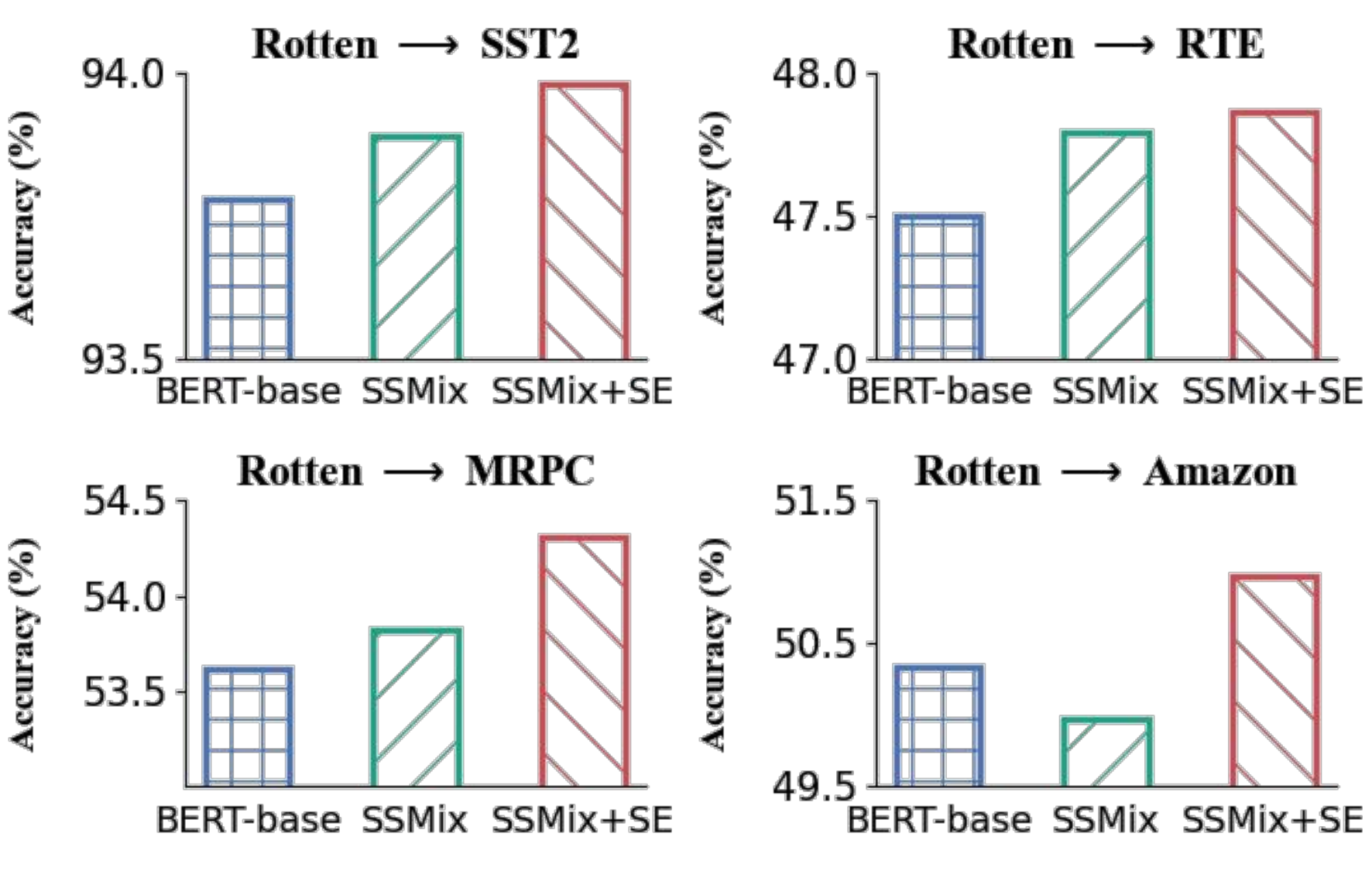}
%[height=4.5cm]表示高度
%[width=9.5cm]表示宽度
%{111.eps}表示eps格式的图片，名为111
\caption{ Analysis of task generalization. The model is fine-tuned on the Rotten tomato task and transferred to four different tasks. We can see that our SE method consistently brings better generalization compared with its  counterparts.}
\label{f3}
\end{figure}

\subsection{Ablation Studies}
% We performed ablation studies to show the effectiveness of each component in self-evolution learning.
We evaluate the impact of each component of our SE, including \textit{i}) learning strategy on mixup, \textit{ii}) instance-specific label smoothing approach, \textit{iii}) coefficient $\alpha$.

\textbf{Impact of Learning Strategy on Mixup.} 
As mentioned in \S\ref{sec:4.3}, we perform the mixup process in an easy-to-hard manner, \textit{i.e.}, first mixing the easy samples and then mixing the hard samples. Here, to investigate the impact of different learning strategies on mixup, we conduct contrastive experiments as following: 1) ``Random'': we randomly select the samples from full dataset; 2) ``Easy-to-hard'': we first train the model with easy samples and then with hard samples; 2) ``Hard-to-easy'': the opposite learning order to ``Easy-to-hard''. The detailed results are listed in Table~\ref{t-5}, and we can find that both ordered learning strategies outperform the baseline ``Random'', indicating the significance of evolution learning. More specifically, ``Easy-to-hard'' achieves the best performance, thus leaving it as the default setting.

% in which we compared the effects of (1) randomly selecting data, (2) selecting data from easy-to-hard, and (3) selecting data from hard-to-easy. We used SSMix as the underlying mixup method and without label smoothing in all cases. The results in Table \ref{t-5} show that selecting data from easy-to-hard, which is the method we used, achieved the best performance. 
% This also indicates that the learning process of the model follows the law of human learning, which progresses from easy to hard.

\textbf{Comparison of Different Label Smoothing.}
A key technology in our method is the instance-specific label smoothing method. To verify its effectiveness, we compare it with vanilla label smoothing and report the results in Table~\ref{t-6}. We show that 1) both label smoothing methods achieve better performance compared to the baseline, confirming the necessity to alleviate the over-confidence problem; 2) our method could further improve the results by a clear margin against vanilla label smoothing. These results prove the effectiveness of our ILS method.
% We investigated the performance of our proposed instance-specific label smoothing compared to the vanilla label smoothing method \cite{szegedy2016rethinking} on the few-shot RTE, MRPC, and SST2 datasets. We report the results in Table \ref{t-6}. We observed that our proposed instance-specific label smoothing method outperforms the general label smoothing method in terms of model performance. This confirms that our proposed label smoothing method is more effective.

% 	\begin{figure}[htbp]%%图,[htbp]是浮动格式
%  \centering   
% 	\subfigure[]{
% 	\label{fig:subfig1}\includegraphics[width=0.4\textwidth]{img_sst2.png}
% }

% 	\subfigure[]{
% 	\label{fig:subfig2}\includegraphics[width=0.4\textwidth]{img_subj.png}
% }	
% 	% \hspace{2mm}
% 	% \subfigure[]{
% 	% 	\includegraphics[width=2.5cm,height=2.5cm]{figures//hx2.png} \label{Fig.6(b)}
% 	% }
	
% 	\caption{Performance (test accuracy (\%)) on SST2(a) and SUBJ (b) with 10 labeled data and varying $\alpha$ set for self-evolution learning. }\label{fig:figure1}
% 	\end {figure}

\begin{figure*}[htbp]
	\centering
 \includegraphics[width=0.85\textwidth]{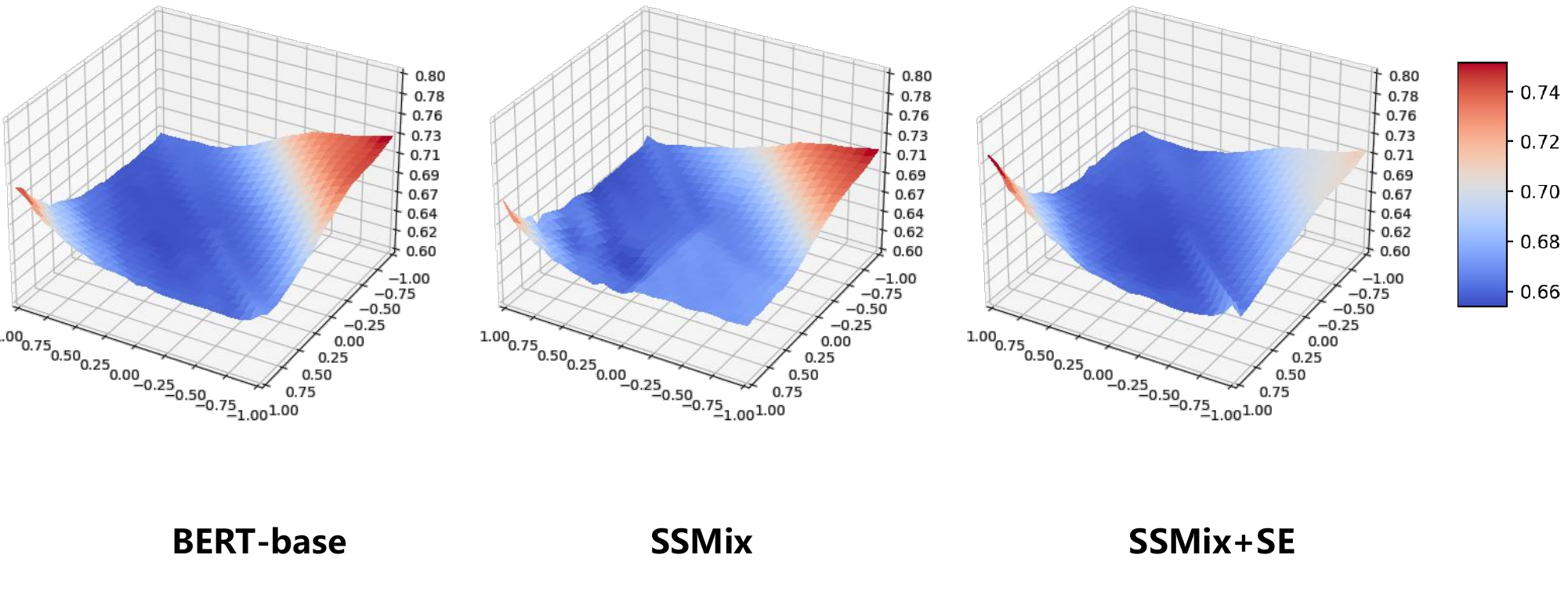}
	\caption{The 3D loss surface comparison between baseline, vanilla SSMix, and our SE methods applied to BERT-base. Note that the PLMs are fine-tuned on the Rotten tomato task. It can be seen that SE methods significantly smooth the loss surface, i.e., improving the model generalization effectively.}\label{fig:loss}
\end{figure*}

\textbf{Impact of Coefficient $\alpha$.}
The weight $\alpha$ in Eq.~\ref{eq:5} is used to control the ratio of label smoothing, which is an important hyper-parameter. In this part, we examine its impact by evaluating the performance with different $\alpha$ on SUBJ task, and illustrate the results in Figure \ref{fig:subfig2}. As shown, compared with the baseline, our method consistently achieves better performance across all ratios of $\alpha$. More specifically, the case of $\alpha = 0.1$ performs best, and we hereby use the setting in our experiments. 

% Figure \ref{fig:figure1} shows the results on different $\alpha$ settings for self-evolution learning. It is evident from the results that the model's performance is generally better when the $\alpha$ is set to 0.1 compared to other settings. Therefore, we conclude that setting the $\alpha$ parameter to 0.1 is suitable for our method.

% \paragraph{Varing the Base Model.} We upgraded BERT-base to BERT-large and conducted tests on various datasets. According to the results presented in Table \ref{t-3}, our method still significantly improved performance. Meanwhile, it is worth noting that our proposed approach demonstrates a more pronounced improvement in BERT-large compared to BERT-base. This can be attributed to the fact that BERT-large has a larger number of parameters, which makes it more susceptible to overconfidence issues when trained on a very small amount of data. Therefore, our approach can have a greater impact in such scenarios.

\subsection{Expanding to High-resource Scenarios}
Although our work mainly focuses on the data augmentation in few-shot tasks, we also investigate whether our method still works in the high-resource scenarios. Specifically, we change the percentage of training data used from 20\% to 100\% and illustrate the results of several tasks in Figure~\ref{f1}. 

% In this study, we investigated the performance of our proposed method with varying amounts of training data. We changed the percentage of training data used from 20\% to 100\% and the results are shown in Figure \ref{f1}. 
As expected, our method achieves significant performance improvements when the amount of training data was extremely limited, continuing to confirm the effectiveness of our method. Moreover, we can also observe performance gains brought by our SE in the other relatively high-resource scenarios. These results prove the universality of our method.

\subsection{Analysis of Model Generalization}
To investigate whether our SE can bring better model generalization, we conduct experiments from two perspectives: \textit{i}) measuring the cross-task zero-shot performance, and \textit{ii}) visualizing the loss landscapes of models.

\textbf{Task Generalization.} 
The performance of out-of-domain (OOD) data is widely used to verify the model generalization~\cite{xu2021raise, zhong2022improving}. Hence, we follow~\citet{zhong2022improving} and and evaluate the performance of models on several OOD data. In practice, we first fine-tune BERT-based models trained with different methods (including ``Baseline'', ``SSMiX'', and ``SSMix+SE'') on the Rotten Tomato task, and then inference on other tasks, \textit{i.e.}, SST2, MRPC, RTE, and Amazon. The results are illustrated in Figure~\ref{f3}.
We observe that ``SSMix+SE'' consistently outperforms the other counterparts. To be more specific, compared with baseline, our SE brings a +0.47 average improvement score on these tasks, indicating that our method boosts the performance of models on OOD data.

% We also discussed the generalization ability of the models trained using our method. We conducted Out-of-distribution Generalization experiments (ood) by training models on the Rotten Tomato dataset using three different methods and evaluating them on other target datasets. The results presented in Figure \ref{f3} show that our method produces models with better performance on multiple tasks. This indicates that the models trained using our method have better generalization ability and robustness.

\textbf{Visualization of Loss Landscape.} 
To have a closer look, we also visualize the loss landscapes of different BERT-base models fine-tuned on the Rotten Tomato task. In practice, we follow the ``filter normalized'' setting in~\citet{li2018visualizing} and show the 3D loss surface results in Figure~\ref{fig:loss}. 
% by sampling 25$\times$25 points in the range of [-1, 1] from random ``filter normalized'' directions, as implemented in~\cite{li2018visualizing}. 
% To gain a more intuitive understanding, we visualize the loss landscape of different fine-tuned BERT-base models on the Rotten Tomato dataset. Specifically, we show the 3D loss surface result in Figure \ref{fig:loss} with filter normalized setting \cite{li2018visualizing}. 
We can see that our method has flatter smoother surfaces compared to others. This result proves that SE can smooth the loss landscape and improve the generalization of models effectively.

\section{Conclusion}
In this paper, we propose a simple-yet-effective self-evolution (SE) learning mechanism to improve the existing mixup methods on text classification tasks.  SE for mixup follows two stages: conducting data division based on the degree of difficulty and mixup based on the order from easy to hard. SE can be used in various mixup methods to generate more adaptive and model-friendly pseudo samples for model training. Also, to avoid over-confidence in the model, we propose a novel instance-specific label smoothing approach. Extensive experiments on four popular mixup methods, EmbedMix, TMix, SSMix, and TreeMix, verify the effectiveness of our method. Quantitative analyses and in-depth discussions show our method improves the generalization, and robustness of models.
\label{sec:bibtex}

% \begin{table*}
% \centering
% \begin{tabular}{c|ccccc}
% \hline
% \textbf{Method} & \textbf{Model} & \textbf{SST2}&\textbf{Rotten tomato} & \textbf{CB} & \textbf{RTE} \\
% \hline
%  % \multirow{4}{*}{AG News}    & VAMPIRE & - &83.9  & 86.2 &\multirow{4}{*}{DBpedia} & VAMPIRE&- &-  &- \\ \cline{2-5}\cline{7-10}58.83	37.49	49.52
% \multirow{2}{*}{Tmix} & \verb|Tmix| & 54.94 & 56.95 & 41.06 & 49.60 \\ \cline{2-6} 

%  &\verb|Tmix+Self-evolution| &\makecell{57.56\\ \textbf{(+2.62)}} & \makecell{58.23\\ \textbf{(+1.28)}} & \makecell{42.85 \\ \textbf{(+1.79)}} & \makecell{49.99\\ \textbf{(+0.39)}} \\
% \hline
% \multirow{2}{*}{Embedmix} & \verb|Embedmix| & 53.11 & 58.83 & 37.49 & 49.52 \\ \cline{2-6} 
% &\verb|Embedmix+Self-evolution| &\makecell{55.89\\ \textbf{(+2.78)}} & \makecell{60.10\\ \textbf{(+1.27)}} & \makecell{41.07\\ \textbf{(+3.58)}} & \makecell{49.88\\ \textbf{(+0.36)}} \\
% % \citeyearpar{ct1965} & \verb|\citeyearpar| & \verb|\shortcite| \\
% % \citeposs{ct1965} & \verb|\citeposs| & no equivalent \\
% % \citep[FFT;][]{ct1965} &  \verb|\citep[FFT;][]| & no equivalent\\
% \hline
% \end{tabular}
% \caption{\label{citation-guide}
% Experimental results of comparison with other mixup method. All values are average accuracy (\%) of five runs with different seeds. Models are trained with 10 labeled data per class. 
% }
% \end{table*}

\section*{Limitations}
Our work has several potential limitations. First, due to limited computational resources, we only validate our self-evolution learning on base- and large-size BERT models. Expanding our experiments to larger model sizes would make our work more convincing. 
On the other hand, for the results of baseline methods, we should compare our results with those in the original paper for a fair comparison. However, due to the difference of PLMs and tasks used in the other baselines and ours, it is unreasonable to compare the results directly. Hence, as an alternative, we only reproduce the results in our settings using the code in the corresponding papers.
% On the other hand, our replication of other unopened sourced work may cause some of its results to be low. 

\section*{Ethics Statement}
We take ethical considerations very seriously, and strictly adhere to the EMNLP Ethics Policy. This paper proposes a self-evolution learning algorithm to improve the existing mixup strategy. The proposed approach aims to precisely augment the few-shot training data with the original training corpus, instead of encouraging the model to generate new sentences that may cause the ethical problem. Moreover, all pre-trained language models and downstream datasets used in this paper are publicly available and have been widely adopted by researchers. Thus, we believe that this research will not pose ethical issues.

 \section*{Acknowledgements}
%Research on this paper was supported by the National Natural Science Foundation of China under Grant No. 22XJ01012.
This work is supported by the following foundations: the National Natural Science Foundation of China under Grant No.62025208 and No.62306330, the Xiangjiang Laboratory Foundation under Grant No.22XJ01012.

\bibliography{anthology,custom}

\begin{thebibliography}{36}
\expandafter\ifx\csname natexlab\endcsname\relax\def\natexlab#1{#1}\fi

\bibitem[{{Bailey} and {Chopra}(2018)}]{2018arXiv180402063B}
Katherine {Bailey} and Sunny {Chopra}. 2018.
\newblock \href {https://doi.org/10.48550/arXiv.1804.02063} {{Few-Shot Text
  Classification with Pre-Trained Word Embeddings and a Human in the Loop}}.
\newblock \emph{arXiv e-prints}, page arXiv:1804.02063.

\bibitem[{Bansal et~al.(2020)Bansal, Jha, and
  McCallum}]{bansal-etal-2020-learning}
Trapit Bansal, Rishikesh Jha, and Andrew McCallum. 2020.
\newblock \href {https://doi.org/10.18653/v1/2020.coling-main.448} {Learning to
  few-shot learn across diverse natural language classification tasks}.
\newblock In \emph{Proceedings of the 28th International Conference on
  Computational Linguistics}, pages 5108--5123, Barcelona, Spain (Online).
  International Committee on Computational Linguistics.

\bibitem[{Brown et~al.(2020)Brown, Mann, Ryder, Subbiah, Kaplan, Dhariwal,
  Neelakantan, Shyam, Sastry, Askell et~al.}]{brown2020language}
Tom Brown, Benjamin Mann, Nick Ryder, Melanie Subbiah, Jared~D Kaplan, Prafulla
  Dhariwal, Arvind Neelakantan, Pranav Shyam, Girish Sastry, Amanda Askell,
  et~al. 2020.
\newblock Language models are few-shot learners.
\newblock \emph{Advances in neural information processing systems},
  33:1877--1901.

\bibitem[{Cao et~al.(2021)Cao, Ding, Tian, and Fang}]{cao2021towards}
Yu~Cao, Liang Ding, Zhiliang Tian, and Meng Fang. 2021.
\newblock Towards efficiently diversifying dialogue generation via embedding
  augmentation.
\newblock In \emph{ICASSP 2021-2021 IEEE International Conference on Acoustics,
  Speech and Signal Processing (ICASSP)}, pages 7443--7447. IEEE.

\bibitem[{Chen et~al.(2020)Chen, Yang, and Yang}]{chen-etal-2020-mixtext}
Jiaao Chen, Zichao Yang, and Diyi Yang. 2020.
\newblock \href {https://doi.org/10.18653/v1/2020.acl-main.194} {{M}ix{T}ext:
  Linguistically-informed interpolation of hidden space for semi-supervised
  text classification}.
\newblock In \emph{Proceedings of the 58th Annual Meeting of the Association
  for Computational Linguistics}, pages 2147--2157, Online. Association for
  Computational Linguistics.

\bibitem[{{Devlin} et~al.(2018){Devlin}, {Chang}, {Lee}, and
  {Toutanova}}]{2018arXiv181004805D}
Jacob {Devlin}, Ming-Wei {Chang}, Kenton {Lee}, and Kristina {Toutanova}. 2018.
\newblock \href {https://doi.org/10.48550/arXiv.1810.04805} {{BERT:
  Pre-training of Deep Bidirectional Transformers for Language Understanding}}.
\newblock \emph{arXiv e-prints}, page arXiv:1810.04805.

\bibitem[{Devlin et~al.(2019)Devlin, Chang, Lee, and
  Toutanova}]{devlin-etal-2019-bert}
Jacob Devlin, Ming-Wei Chang, Kenton Lee, and Kristina Toutanova. 2019.
\newblock \href {https://doi.org/10.18653/v1/N19-1423} {{BERT}: Pre-training of
  deep bidirectional transformers for language understanding}.
\newblock In \emph{Proceedings of the 2019 Conference of the North {A}merican
  Chapter of the Association for Computational Linguistics: Human Language
  Technologies, Volume 1 (Long and Short Papers)}, pages 4171--4186,
  Minneapolis, Minnesota. Association for Computational Linguistics.

\bibitem[{Fei-Fei et~al.(2006)Fei-Fei, Fergus, and Perona}]{fei2006one}
Li~Fei-Fei, Robert Fergus, and Pietro Perona. 2006.
\newblock One-shot learning of object categories.
\newblock \emph{IEEE transactions on pattern analysis and machine
  intelligence}, 28(4):594--611.

\bibitem[{Geng et~al.(2020)Geng, Li, Li, Sun, and Zhu}]{geng-etal-2020-dynamic}
Ruiying Geng, Binhua Li, Yongbin Li, Jian Sun, and Xiaodan Zhu. 2020.
\newblock \href {https://doi.org/10.18653/v1/2020.acl-main.102} {Dynamic memory
  induction networks for few-shot text classification}.
\newblock In \emph{Proceedings of the 58th Annual Meeting of the Association
  for Computational Linguistics}, pages 1087--1094, Online. Association for
  Computational Linguistics.

\bibitem[{Guo et~al.(2019)Guo, Mao, and Zhang}]{guo2019augmenting}
Hongyu Guo, Yongyi Mao, and Richong Zhang. 2019.
\newblock Augmenting data with mixup for sentence classification: An empirical
  study.
\newblock \emph{arXiv preprint arXiv:1905.08941}.

\bibitem[{Kobayashi(2018)}]{kobayashi2018contextual}
Sosuke Kobayashi. 2018.
\newblock Contextual augmentation: Data augmentation by words with paradigmatic
  relations.
\newblock \emph{arXiv preprint arXiv:1805.06201}.

\bibitem[{Koco{\'n} et~al.(2023)Koco{\'n}, Cichecki, Kaszyca, Kochanek,
  Szyd{\l}o, Baran, Bielaniewicz, Gruza, Janz, Kanclerz
  et~al.}]{kocon2023chatgpt}
Jan Koco{\'n}, Igor Cichecki, Oliwier Kaszyca, Mateusz Kochanek, Dominika
  Szyd{\l}o, Joanna Baran, Julita Bielaniewicz, Marcin Gruza, Arkadiusz Janz,
  Kamil Kanclerz, et~al. 2023.
\newblock Chatgpt: Jack of all trades, master of none.
\newblock \emph{Information Fusion}, page 101861.

\bibitem[{Li et~al.(2018)Li, Xu, Taylor, Studer, and
  Goldstein}]{li2018visualizing}
Hao Li, Zheng Xu, Gavin Taylor, Christoph Studer, and Tom Goldstein. 2018.
\newblock Visualizing the loss landscape of neural nets.
\newblock \emph{Advances in neural information processing systems}, 31.

\bibitem[{Loshchilov and Hutter(2018)}]{loshchilov2018decoupled}
Ilya Loshchilov and Frank Hutter. 2018.
\newblock Decoupled weight decay regularization.
\newblock In \emph{ICLR}.

\bibitem[{Lu et~al.(2023{\natexlab{a}})Lu, Huang, Tian, Zhao, Fei, and
  Li}]{lu2023meta}
Menglong Lu, Zhen Huang, Zhiliang Tian, Yunxiang Zhao, Xuanyu Fei, and
  Dongsheng Li. 2023{\natexlab{a}}.
\newblock Meta-tsallis-entropy minimization: A new self-training approach for
  domain adaptation on text classification.
\newblock \emph{arXiv preprint arXiv:2308.02746}.

\bibitem[{Lu et~al.(2023{\natexlab{b}})Lu, Huang, Zhao, Tian, Liu, and
  Li}]{lu-etal-2023-damstf}
Menglong Lu, Zhen Huang, Yunxiang Zhao, Zhiliang Tian, Yang Liu, and Dongsheng
  Li. 2023{\natexlab{b}}.
\newblock \href {https://doi.org/10.18653/v1/2023.acl-long.92} {{D}a{MSTF}:
  Domain adversarial learning enhanced meta self-training for domain
  adaptation}.
\newblock In \emph{Proceedings of the 61st Annual Meeting of the Association
  for Computational Linguistics (Volume 1: Long Papers)}, pages 1650--1668,
  Toronto, Canada. Association for Computational Linguistics.

\bibitem[{Lu et~al.(2023{\natexlab{c}})Lu, Qiu, Ding, Xie, and
  Tao}]{lu2023error}
Qingyu Lu, Baopu Qiu, Liang Ding, Liping Xie, and Dacheng Tao.
  2023{\natexlab{c}}.
\newblock Error analysis prompting enables human-like translation evaluation in
  large language models: A case study on chatgpt.
\newblock \emph{arXiv preprint arXiv:2303.13809}.

\bibitem[{Ouyang et~al.(2022)Ouyang, Wu, Jiang, Almeida, Wainwright, Mishkin,
  Zhang, Agarwal, Slama, Ray et~al.}]{ouyang2022training}
Long Ouyang, Jeffrey Wu, Xu~Jiang, Diogo Almeida, Carroll Wainwright, Pamela
  Mishkin, Chong Zhang, Sandhini Agarwal, Katarina Slama, Alex Ray, et~al.
  2022.
\newblock Training language models to follow instructions with human feedback.
\newblock \emph{Advances in Neural Information Processing Systems},
  35:27730--27744.

\bibitem[{Park and Caragea(2022)}]{park-caragea-2022-calibration}
Seo~Yeon Park and Cornelia Caragea. 2022.
\newblock \href {https://doi.org/10.18653/v1/2022.acl-long.368} {On the
  calibration of pre-trained language models using mixup guided by area under
  the margin and saliency}.
\newblock In \emph{Proceedings of the 60th Annual Meeting of the Association
  for Computational Linguistics (Volume 1: Long Papers)}, pages 5364--5374,
  Dublin, Ireland. Association for Computational Linguistics.

\bibitem[{Peng et~al.(2023)Peng, Ding, Zhong, Shen, Liu, Zhang, Ouyang, and
  Tao}]{peng2023towards}
Keqin Peng, Liang Ding, Qihuang Zhong, Li~Shen, Xuebo Liu, Min Zhang, Yuanxin
  Ouyang, and Dacheng Tao. 2023.
\newblock Towards making the most of chatgpt for machine translation.
\newblock \emph{arXiv preprint arXiv:2303.13780}.

\bibitem[{Sawhney et~al.(2022)Sawhney, Thakkar, Pandit, Soun, Jin, Yang, and
  Flek}]{sawhney-etal-2022-dmix}
Ramit Sawhney, Megh Thakkar, Shrey Pandit, Ritesh Soun, Di~Jin, Diyi Yang, and
  Lucie Flek. 2022.
\newblock \href {https://doi.org/10.18653/v1/2022.acl-short.67} {{DM}ix:
  Adaptive distance-aware interpolative mixup}.
\newblock In \emph{Proceedings of the 60th Annual Meeting of the Association
  for Computational Linguistics (Volume 2: Short Papers)}, pages 606--612,
  Dublin, Ireland. Association for Computational Linguistics.

\bibitem[{Shleifer(2019)}]{shleifer2019low}
Sam Shleifer. 2019.
\newblock Low resource text classification with ulmfit and backtranslation.
\newblock \emph{arXiv preprint arXiv:1903.09244}.

\bibitem[{Sun et~al.(2020)Sun, Xia, Yin, Liang, Yu, and He}]{sun2020mixup}
Lichao Sun, Congying Xia, Wenpeng Yin, Tingting Liang, Philip~S Yu, and Lifang
  He. 2020.
\newblock Mixup-transformer: dynamic data augmentation for nlp tasks.
\newblock \emph{arXiv preprint arXiv:2010.02394}.

\bibitem[{Szegedy et~al.(2016)Szegedy, Vanhoucke, Ioffe, Shlens, and
  Wojna}]{szegedy2016rethinking}
Christian Szegedy, Vincent Vanhoucke, Sergey Ioffe, Jon Shlens, and Zbigniew
  Wojna. 2016.
\newblock Rethinking the inception architecture for computer vision.
\newblock In \emph{Proceedings of the IEEE conference on computer vision and
  pattern recognition}, pages 2818--2826.

\bibitem[{Verma et~al.(2019)Verma, Lamb, Beckham, Najafi, Mitliagkas,
  Lopez-Paz, and Bengio}]{verma2019manifold}
Vikas Verma, Alex Lamb, Christopher Beckham, Amir Najafi, Ioannis Mitliagkas,
  David Lopez-Paz, and Yoshua Bengio. 2019.
\newblock Manifold mixup: Better representations by interpolating hidden
  states.
\newblock In \emph{International conference on machine learning}, pages
  6438--6447. PMLR.

\bibitem[{Wang et~al.(2019)Wang, Pruksachatkun, Nangia, Singh, Michael, Hill,
  Levy, and Bowman}]{wang2019superglue}
Alex Wang, Yada Pruksachatkun, Nikita Nangia, Amanpreet Singh, Julian Michael,
  Felix Hill, Omer Levy, and Samuel Bowman. 2019.
\newblock Superglue: A stickier benchmark for general-purpose language
  understanding systems.
\newblock In \emph{NeurIPS}.

\bibitem[{Wang et~al.(2018)Wang, Singh, Michael, Hill, Levy, and
  Bowman}]{wang2018glue}
Alex Wang, Amanpreet Singh, Julian Michael, Felix Hill, Omer Levy, and Samuel
  Bowman. 2018.
\newblock Glue: A multi-task benchmark and analysis platform for natural
  language understanding.
\newblock In \emph{EMNLP}.

\bibitem[{Wei and Zou(2019)}]{wei2019eda}
Jason Wei and Kai Zou. 2019.
\newblock Eda: Easy data augmentation techniques for boosting performance on
  text classification tasks.
\newblock \emph{arXiv preprint arXiv:1901.11196}.

\bibitem[{Wu et~al.(2019)Wu, Lv, Zang, Han, and Hu}]{wu2019conditional}
Xing Wu, Shangwen Lv, Liangjun Zang, Jizhong Han, and Songlin Hu. 2019.
\newblock Conditional bert contextual augmentation.
\newblock In \emph{Computational Science--ICCS 2019: 19th International
  Conference, Faro, Portugal, June 12--14, 2019, Proceedings, Part IV 19},
  pages 84--95. Springer.

\bibitem[{Xu et~al.(2021)Xu, Luo, Zhang, Tan, Chang, Huang, and
  Huang}]{xu2021raise}
Runxin Xu, Fuli Luo, Zhiyuan Zhang, Chuanqi Tan, Baobao Chang, Songfang Huang,
  and Fei Huang. 2021.
\newblock Raise a child in large language model: Towards effective and
  generalizable fine-tuning.
\newblock In \emph{EMNLP}.

\bibitem[{Yoon et~al.(2021)Yoon, Kim, and Park}]{yoon2021ssmix}
Soyoung Yoon, Gyuwan Kim, and Kyumin Park. 2021.
\newblock Ssmix: Saliency-based span mixup for text classification.
\newblock \emph{arXiv preprint arXiv:2106.08062}.

\bibitem[{Yu et~al.(2018)Yu, Guo, Yi, Chang, Potdar, Cheng, Tesauro, Wang, and
  Zhou}]{yu-etal-2018-diverse}
Mo~Yu, Xiaoxiao Guo, Jinfeng Yi, Shiyu Chang, Saloni Potdar, Yu~Cheng, Gerald
  Tesauro, Haoyu Wang, and Bowen Zhou. 2018.
\newblock \href {https://doi.org/10.18653/v1/N18-1109} {Diverse few-shot text
  classification with multiple metrics}.
\newblock In \emph{Proceedings of the 2018 Conference of the North {A}merican
  Chapter of the Association for Computational Linguistics: Human Language
  Technologies, Volume 1 (Long Papers)}, pages 1206--1215, New Orleans,
  Louisiana. Association for Computational Linguistics.

\bibitem[{Yun et~al.(2019)Yun, Han, Oh, Chun, Choe, and Yoo}]{yun2019cutmix}
Sangdoo Yun, Dongyoon Han, Seong~Joon Oh, Sanghyuk Chun, Junsuk Choe, and
  Youngjoon Yoo. 2019.
\newblock Cutmix: Regularization strategy to train strong classifiers with
  localizable features.
\newblock In \emph{Proceedings of the IEEE/CVF international conference on
  computer vision}, pages 6023--6032.

\bibitem[{Zhang et~al.(2022)Zhang, Yang, and Yang}]{zhang-etal-2022-treemix}
Le~Zhang, Zichao Yang, and Diyi Yang. 2022.
\newblock \href {https://doi.org/10.18653/v1/2022.naacl-main.385} {{T}ree{M}ix:
  Compositional constituency-based data augmentation for natural language
  understanding}.
\newblock In \emph{Proceedings of the 2022 Conference of the North American
  Chapter of the Association for Computational Linguistics: Human Language
  Technologies}, pages 5243--5258, Seattle, United States. Association for
  Computational Linguistics.

\bibitem[{Zhong et~al.(2023)Zhong, Ding, Liu, Du, and Tao}]{zhong2023chat}
Qihuang Zhong, Liang Ding, Juhua Liu, Bo~Du, and Dacheng Tao. 2023.
\newblock \href {https://arxiv.org/abs/2302.10198} {Can chatgpt understand too?
  a comparative study on chatgpt and fine-tuned bert}.
\newblock \emph{arXiv preprint}.

\bibitem[{Zhong et~al.(2022)Zhong, Ding, Shen, Mi, Liu, Du, and
  Tao}]{zhong2022improving}
Qihuang Zhong, Liang Ding, Li~Shen, Peng Mi, Juhua Liu, Bo~Du, and Dacheng Tao.
  2022.
\newblock Improving sharpness-aware minimization with fisher mask for better
  generalization on language models.
\newblock In \emph{Findings of EMNLP}.

\end{thebibliography}
\bibliographystyle{acl_natbib}

\appendix

\section{Appendix}

\label{sec:appendix}
\subsection{Details of Datasets}
\label{appendix_data}
\begin{table}[!h]
    \centering
    \begin{threeparttable}
        \resizebox{0.48\textwidth}{!}{
        \begin{tabular}{lccc}
            \toprule
            \multicolumn{1}{c}{Dataset} & Task & $\#$ Label & Size  \\
            \midrule
            SST-2 & Sentiment & 2 & 67k / 1.8k \\
            \midrule
            RTE & NLI & 2 & 2.5k / 3k  \\
            \midrule
            MRPC & Paraphrase & 2 & 3.7k / 1.7k  \\
            \midrule
            CB & NLI & 3 & 556 / 250  \\
            \midrule
            SUBJ & Classification & 2 & 8k / 2k  \\
                        \midrule
            Rotten tomato & Sentiment & 2 & 8.53k / 1.07k  \\
                        \midrule
            Amazon counterfactual & Classification & 2 & 5k / 5k  \\
            % \midrule
            % TREC-coarse & Classification & 6 & 5.5k / 500  \\
            % \midrule
            % TREC-fine & Classification & 47 & 5.5k / 500  \\
            \bottomrule
        \end{tabular}}
    \end{threeparttable}
    \caption{Dataset name, task, number of total labels, and dataset size of datasets we used as a benchmark. The task column describes the objective of each dataset.  }
    \label{tab:dataset_summary}
\end{table}

\subsection{Parameter Analysis on $\lambda$}
\begin{table}
\centering
\scalebox{0.65}{
\begin{tabular}{lcccc}
\toprule
\textbf{$\lambda$} & \textbf{0.1} & \textbf{0.2} & \textbf{0.3} & \textbf{0.4}\\
\midrule
TMix & 55.01 & 54.94 & 54.46 &54.93\\
% TMix+Ours. & \textbf{55.66} & \textbf{61.17} & \textbf{65.37} & \textbf{60.78}\\
TMix+Ours. &  55.66(+0.65) & 57.56(+2.62) & 54.93(+0.47) & 55.40(+0.47)\\
\bottomrule
\end{tabular}}
\caption{\label{t-l}
The experimental results were an average of 5 runs with random seeds. The content in the brackets is the improvement of our method. 
}
\end{table}

As stated in Sec\ref{sec:4.2}, We use the parameter $\lambda$ to control the mixing ratio of two samples. Here, we analyze
the influence of different $\lambda$ in detail. In practice, we used TMix as the baseline and conducted experiments with $\lambda$ varying from 0.1 to 0.4. The results are presented in the table \ref{t-l}. As seen, our method brings consistent performance gains across various, indicating that our method is not sensitive to the value of $\lambda$. Notably, "$\lambda$=0.2" achieves the best performance, thus leaving as the default setting.

\end{document}